\documentclass[11pt]{article}
\usepackage{acl2012}
\usepackage{times}
\usepackage{latexsym}
\usepackage{amsmath}
\usepackage{multirow}
\usepackage{url}
\usepackage{acronym}
\usepackage{xspace}
\usepackage{color}
\usepackage{algorithm}
\usepackage{algorithmic}
\usepackage{verbatim} 
\usepackage{graphicx}
\usepackage{footmisc}
\usepackage{multirow}
\usepackage{amssymb}

\acrodef{QA}{Question-Answering}
\acrodef{AE}{Answer Extraction}
\acrodef{PR}{Passage Retrieval}
\acrodef{QG}{Question Generation}
\acrodef{NER}{Named Entity Recognition}
\acrodef{Q/A}{Question/Answer}
\acrodef{WWW}{World Wide Web}
\acrodef{IE}{Information Extraction}
\acrodef{OIE}{Open Information Extraction}
\acrodef{PoS}{Part-of-Speech}
\acrodef{TREC}{Text REtrieval Conference}
\acrodef{NYT}{New York Times}
\acrodef{KB}{Knowledge Base}
\acrodef{QI}{Question Processing}

\newcommand{\ja}{\textsc{Just.Ask}\xspace}
\newcommand{\q}[1]{\textit{#1}}
\newcommand{\qum}[1]{\textit{#1}}
\renewcommand{\a}[1]{\textit{#1}}
\newcommand{\s}[1]{\textit{#1}}
\newcommand{\p}[1]{{``\small #1''}}

\newcommand{\qa}[2]{(\qum{#1}, \a{#2})}

\newcommand\eg{{\it e.g.}\xspace}
\newcommand\etal{{\it et al.}\xspace}

\title{Learning to answer questions}

\author{Ana Cristina Mendes \\
  L$^2$F INESC-ID Lisboa \\
  Instituto Superior Tecnico\\
   \And
  Lu\'{i}sa Coheur\\
 L$^2$F INESC-ID Lisboa \\
Instituto Superior Tecnico\\
\And
  S\'{e}rgio Curto \\
  L$^2$F INESC-ID Lisboa \\
}

\date{}

\begin{document}
\maketitle
\begin{abstract}

We present an open-domain \acl{QA} system that learns to answer questions based on successful past interactions. We follow a pattern-based approach to \acl{AE}, where (lexico-syntactic) patterns that relate a question to its answer are automatically learned and used to answer future questions. 
Results show that our approach contributes to the system's best performance when it is conjugated with typical \acl{AE} strategies. Moreover, it allows the system to learn with the answered questions and to rectify wrong or unsolved past questions. 

\end{abstract}

\section{Introduction}

\ac{QA} systems aim at delivering correct answers to questions posed in natural language. In this paper, we present our approach to open-domain \ac{QA}, motivated by the hypothesis that if a system is given feedback about the correctness of its answers, it can learn to answer new questions. 

Our approach relies on iteratively learning to answer new questions based on previously answered questions. In the basis of this work, there is a pattern-based strategy to \ac{AE}, with (lexico-syntactic) patterns learned from pairs composed of natural language questions and their answers. Thus, when a new question is posed to the system, the applicable patterns are unified with the sentences where the new answers might lie. If the unification succeeds, possible answers are extracted; if not, the unification is relaxed. If feedback is given to the system about the correctness of its answers, the posed question and the returned correct answer are used to feed a new pattern learning cycle. 

In the evaluation we compare our approach with classical \ac{AE} strategies. Results show that our approach contributes to the system's best achieved performance when it is conjugated with the remaining \ac{AE} strategies. Moreover, our approach allows the system to learn with posed questions and is able to rectify past questions.

This paper is organized as follows: Section~\ref{sec:rw} refers to related work; Section~\ref{sec:qa} describes the approach; Section~\ref{sec:eval} presents the evaluation and Section~\ref{sec:concl} highlights the main conclusions and points to future work.

\section{Related Work}\label{sec:rw}

The pattern-based approach to \ac{AE} has received much attention in \ac{QA}, specially after the TREC-10 conference, where the winning system~\cite{Soubbotin01} relied on a list of hand-built surface patterns.

The focus has turned to automatic pattern learning, with many approaches taking advantage of the large amount of information on the Web. For example, \cite{ravichandran02} and \cite{zhang02} learn surface patterns for a particular relation. Javelin~\cite{shima10} adds flexibility to the learned patterns by allowing, for instance, certain terms to be generalized into named entities. In all the previous cases, seeds are sets of two (manually chosen) entities. On the contrary, Ephyra system~\cite{schlaefer06} takes as input questions and answers, and learns patterns between the answer and two or more key-phrases. However, these are extracted from the question by hand-made patterns.
Previous approaches learn lexical patterns; others learn patterns also conveying syntactic information in the form of dependency relations~\cite{shen05,bouma11}. Particularly, Shen \etal define a pattern as the smallest dependency tree that conveys the answer and one question term.
 


\section{From Answered Questions to QA}\label{sec:qa}

\subsection{\ja~-- An Open Domain \ac{QA} System}\label{sec:ja}

\ja is the open-domain \ac{QA} system that we will use in all our experiments. It follows the typical pipelined architecture of \ac{QA} systems composed of three main stages: \acl{QI}, \acl{PR} and \ac{AE}. 

In the \acl{QI} stage, the Berkeley Parser~\cite{petrov07}, trained on the QuestionBank~\cite{judge06}, analyzes each question. A machine learning-based classifier (SVM) fed with features derived from a rule-based classifier~\cite{silva11} classifies the questions according to Li and Roth's taxonomy~\shortcite{li02}. Concerning the \acl{PR} stage, the relevant passages are retrieved from the information sources (local or the Web). Finally, regarding the \ac{AE} stage, \ja relies on a panoply of strategies and resources, namely: matching of regular expressions, machine learning \ac{NER} (by using the Stanford's Conditional Random Field-based recognizer~\cite{finkel05}), gazetteers, dictionaries constructed \textit{on the fly} and the linguistic patterns previously learned. We will call NER to the combination of all the implemented \ac{AE} strategies, except the pattern-based strategy.

\subsection{Pattern Learning in a Nutshell}

To learn patterns, we follow the approach described in \cite{omitted}. Briefly, lexico-syntactic patterns are learned from \ac{Q/A} pairs (each composed of a question and its answer) and from sentences retrieved from the information sources that contain constituents of these pairs. Each pattern relates a question to its answer, through the constituents of a sentence that contains parts of the question and the answer.  For example, the pattern \p{\textsc{np${_{answer}}$} has \textsc{vbn} \textsc{np}}\footnote{Note that the pattern \p{\textsc{np${_{answer}}$} has \textsc{vbn} \textsc{np}} indicates that the answer is in the first \textsc{np}.} is learned from  the sentence \s{Dante has written The Divine Comedy} and the \ac{Q/A} pair \qa{Who wrote The Divine Comedy?}{Dante}.

All the learned patterns and the source \ac{Q/A} pairs are stored in a \ac{KB}. 

\subsection{Pattern-based \acl{AE}}

\begin{figure}[htb]
\centering
\includegraphics[width=.43\textwidth]{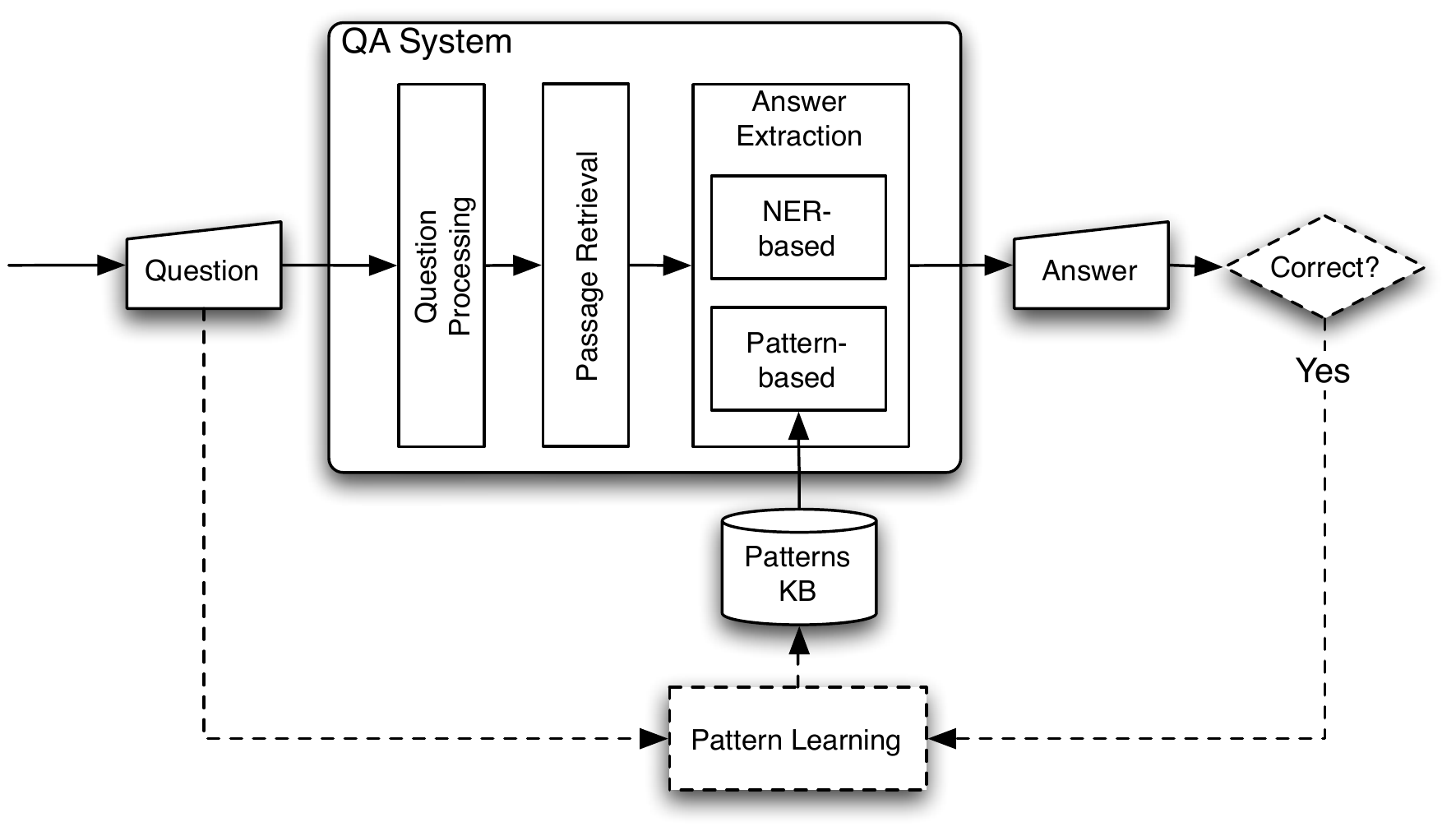}
\caption{Learning from answered questions in \ja.}
\label{img:arch}
\end{figure}

Figure~\ref{img:arch} depicts the flow of information of the pattern-based \ac{AE} strategy with feedback loop. 

Given a new question, its syntactic analysis (performed in the \acl{QI} stage) allows to retrieve from the \ac{KB} all the applicable patterns. Relevant passages are returned in the \acl{PR} stage and, in the \ac{AE} stage, there is a pattern/sentence unification to extract the candidate answers: every sentence in the passages is analyzed and its parse tree is explored in a top-down, left-to-right, depth-first search, unifying the obtained sentence tree components with the lexico-syntactic information in the patterns.

Since the pattern-learning algorithm receives a \ac{Q/A} pair as input, if the system is able to verify the correctness of its answers, the process to learn new patterns is straightforward. Therefore, if positive feedback is given, \ja learns new patterns with previously answered questions.

\subsection{Lexico-Syntactic Relaxation Strategies}

Due to the variability of natural language, a pure pattern/sentence unification is rare. Therefore, the unification process relies in the following:
\begin{enumerate}
\item If there are no patterns in the \ac{KB} that apply to the question -- if the system has not yet seen a question with the same syntactic structure and semantic category of the new question:  \ja runs the \ac{NER} strategies.
\item If there are patterns in the \ac{KB} that apply to the question, but the unification does not succeed and no answer is captured, \ja uses two relaxation strategies: 
a) A lexical distance (Levenshtein, overlap and Jaccard) is permitted between the sentence and the question lexical components. This allows, for instance, to cope with orthographic errors. 
b) A syntactic match is consented between (syntactic) classes that relate with each other as they belong to the same (supper) class (\eg, both \p{\textsc{np}} (proper noun) and \p{\textsc{nns}} (plural noun) are nouns). This allows the pattern \p{\textsc{np} has \textsc{vbn} \textsc{np}} to match a sentence analyzed as \p{\textsc{nn} has \textsc{vbn} \textsc{np}}.
\end{enumerate}

\section{Evaluation}\label{sec:eval}

\subsection{Experimental Setup}

The evaluation is made with a corpus of questions and their correct answers, built from the freely available data of the TREC \ac{QA} tracks, years 2002 to 2007, where the answers were given by the competing systems and judged as correct by human assessors. From the original corpus, we collected the questions whose answers can be found in the \ac{NYT} and discarded those with pronominal anaphora and ellipsis. The resulting corpus contains 1000 \ac{Q/A} pairs. Lucene\footnote{\url{http://lucene.apache.org}} was used to index and retrieve the paragraphs from the \ac{NYT} from 1987 to 2007.

We assume that the selection of the final answer is flawless: if \ja is able to find a correct answer, it will retrieve it. 
To simulate the feedback loop, the answers of the corpus are used: if \ja finds an answer from the corpus, the learning process begins, as a (human) positive feedback would do.

Performance is measured using precision, recall and F-measure~\cite{jurafsky00}. Results are calculated as a function of the number of questions previously evaluated (when evaluating the $i^{th}$ question $P_i$, $R_i$ and $F_i$ are determined).



In a first experiment, we test our approach in 4 scenarios, where we vary the strategies employed for \ac{AE}: 
1) \textit{NER}: \ja runs with the NER strategies; no (pattern) learning process is involved;
2) \textit{Patterns, with reference as fallback}: the pattern-based strategy is used. If no correct answer is found, one answer from the corpus is chosen and used (with the posed question) as input to a new pattern learning cycle. By doing so, we mimic the behavior of a tutor, which explicitly provides the correct answer to the system;\footnote{In this situation, the answer is accounted as wrong.\label{fn:repeat}}
3) \textit{NER and Patterns}: combination of all implemented \ac{AE} strategies;
4) \textit{NER and Patterns, with reference as fallback}: combination of all implemented \ac{AE} strategies, but if no answer is found, an answer from the corpus is used.\footref{fn:repeat}

In a second experiment, we allow \ja to revise wrong or unsolved past questions and try to solve them using recent knowledge. Thus, we stop the interactions at the $(i * n)^{th}$ question ($i=\{100,250,500\}$, $n \in \{1, ..., 10\}$ and $i * n < 1000$) and try to answer all previous wrong or unsolved questions, by using the new learnt patterns (except the ones learnt with the question in revision).


\subsection{Results}



Concerning the first experiment, results at the 1000$^{th}$ question are in Table~\ref{tab:resultsat1000}.\footnote{In other runs, where we shuffled the corpus, we verified similar results and tendencies.}

\begin{table}[ht!]
\centering
\begin{tabular}{l|rrrr|}
\cline{2-5}
& \multicolumn{4}{c|}{\textbf{Scenario}}\\
& \multicolumn{1}{c}{1} & \multicolumn{1}{c}{2} & \multicolumn{1}{c}{3} & \multicolumn{1}{c|}{4}\\
\hline
\hline
$R_{1000}$ & 45.1\%& 12.7\%& 46.4\%& \textbf{47.5\%}\\
$P_{1000}$ & 79.2\%&  43.5\%& \textbf{79.9\%} & 77.2\%\\
$F_{1000}$ & 57.5\%&  19.7\%& 57.8\% & \textbf{58.8\%}\\
\hline
\end{tabular}
\caption{Results of the first experiment.}
\label{tab:resultsat1000}
\end{table}

Table~\ref{tab:resultsat1000i} reports the results of the second experiment at the 1000$^{th}$ question.

\begin{table}[t!h!]
\centering
\begin{tabular}{l|r|rrr|}
\cline{2-5}
& \multicolumn{1}{c|}{\textbf{No}} & \multicolumn{3}{c|}{\textbf{i}}\\
& \multicolumn{1}{c|}{Reiteration} & \multicolumn{1}{c}{100} & \multicolumn{1}{c}{250} & \multicolumn{1}{c|}{500}\\
\hline
\hline
$R_{1000}$ & 12.7\%& \textbf{17.7\%} & 16.8\%& 15.2\%\\
$P_{1000}$ & 43.5\%& \textbf{47.8\%} & 46.3\%& 45.4\%\\
$F_{1000}$ & 19.7\%& \textbf{25.8\%} & 24.7\%& 22.8\%\\
\hline
\end{tabular}
\caption{Results of the second experiment.}
\label{tab:resultsat1000i}
\end{table}

Considering the first experiment, best results (recall and F-measure) are achieved when the pattern based approach is conjugated with \ac{NER}. Precision attains best results in scenario three.

Considering the second experiment, \ja's recall increases when $i$ decreases: the more backward loops, the more correct answers are extracted. We should mention that, for instance, when \ja stops at the 500$^{th}$ question, it correctly answers 25 questions that it was not able to solve before.
Moreover, we verify that if all patterns learned with the 1000 questions are employed, \ja can solve 188 of the total 1000 questions (again the patterns learnt from a question are not used to answer it).

\begin{figure}[tb]
\begin{center}
\includegraphics[width=.42\textwidth]{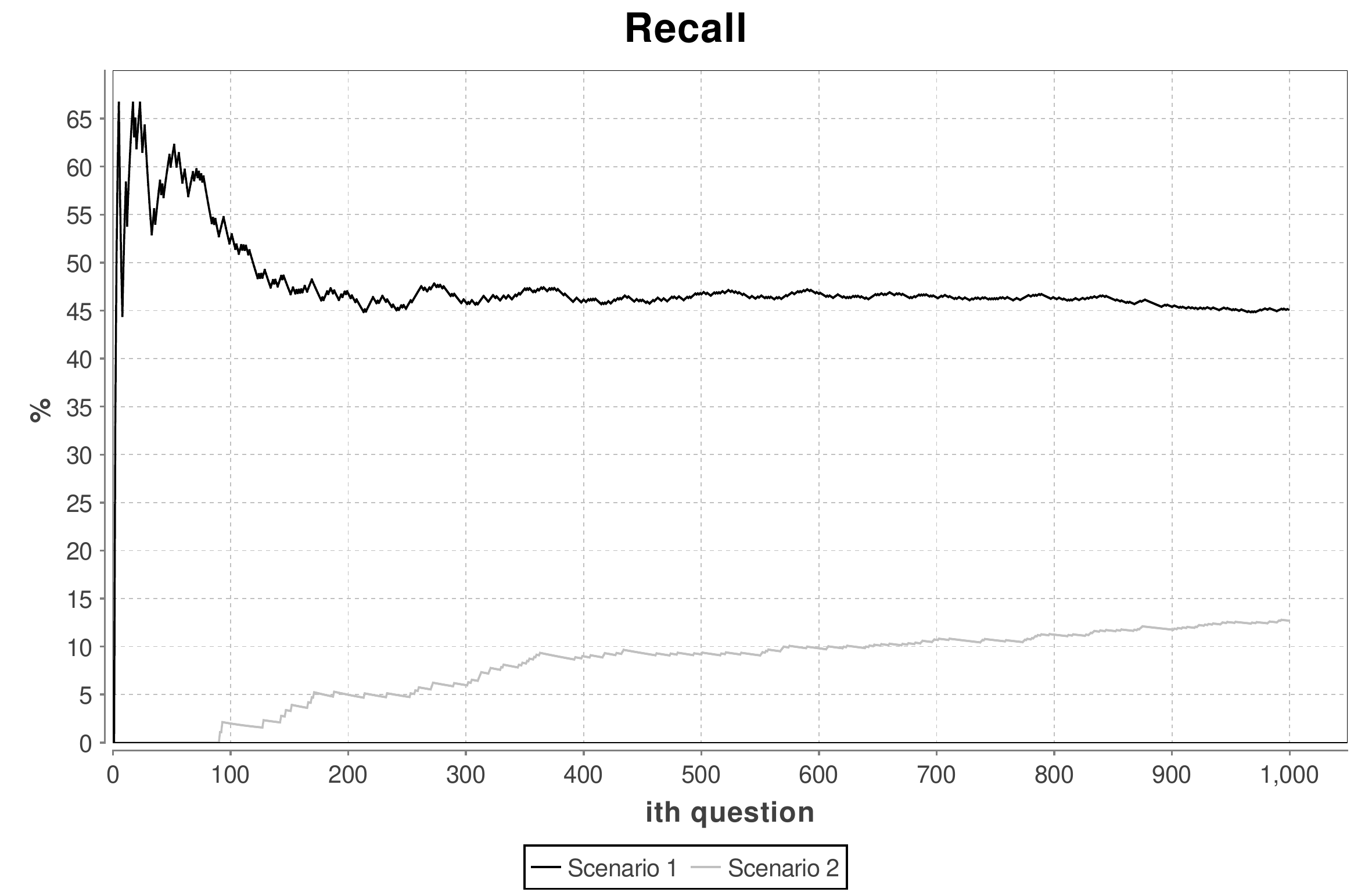}
\caption{Evolution of recall with the posed questions}
\label{tab:recall}
\end{center}
\end{figure}

Figure~\ref{tab:recall} allows us to compare the recall evolution on scenarios one and two. It can be seen that, although leading to the worst results, there is a clear upward tendency of the recall in the second scenario, which is not verified in the scenario where NER is used. Moreover, from the same figure, we can see that, in the second scenario, the recall of \ja starts to increase around the 90$^{th}$ question and keeps on rising with the number of posed questions. This is because the system learns new patterns and correctly answers more questions. 

We should add that we observed much overlap between the remaining scenarios (not shown on the figure). However, the recall becomes slightly higher with the number of answered questions on the third and fourth scenarios when compared with the first one, where \ja is not learning.


Finally, regarding the relaxation strategies, 105 of the 127 correct answers were extracted using an exact unification, as 22 questions were correctly answered when using the relaxation strategies. 

\subsection{Analysis}

Looking at the achieved results, the first fact that needs to be mentioned is that, although we evaluate our system with 1000 questions, only 791 were answerable since Lucene was able to retrieve passages containing the answers for 791 questions only.

Moreover, we find out that patterns aid \ja's performance by discovering answers to misclassified questions. For instance, the questions \q{Who did France beat for the World Cup?} expects a country as an answer. However, since it is classified as requiring the name of an individual, the correct answer \a{Brazil} is not extracted by the \ac{NER} strategies. Patterns were successful in this situation and also to find answers to questions like \q{What is the ``Playboy'' logo?} and \q{How did Malcolm X die?}, for which \ja does not have a specific NER strategy.

To conclude, we believe that the diversity of the questions posed to the system and the small amount of information used to extract answers lead to the significantly poor results of the pattern-based strategy. However, results show increasing performance along the learning process, suggesting that this approach would benefit from more seeds and broader information sources.
Due to the lack of available resources that make this experiment reproducible (either questions and answers, and information sources where the answers are stated), we could not empirically verify the upper bound of recall. 


\section{Conclusions and Future Work}\label{sec:concl}

In this paper, we proposed a fully-automatic pattern-based approach to \ac{AE}, in which the system is itself the provider of the instances (question/answer) that feed the pattern learning algorithm. After new patterns are learned, they are immediately available to extract new answers. 
To the best of our knowledge, \ja is the first system able to learn from successfully answered questions as well as to revise and amend previous questions with newly acquired knowledge.

As future work, we want to learn to score patterns, rewarding the ones that extract correct answers and penalizing those that do not extract any correct candidate answer. Another planned task is to implement other relaxation strategies in the pattern unification.


\bibliographystyle{acl2012}
\bibliography{acl2012bib}

\end{document}